\documentclass[letterpaper, 10 pt, conference]{ieeeconf}  

\IEEEoverridecommandlockouts                              
\usepackage{graphicx}
\usepackage{siunitx}
\usepackage[utf8]{inputenc}
\usepackage[T1]{fontenc}
\usepackage{amsfonts}
\newcommand{\fig}[1]{Fig.~\ref{#1}}

\usepackage{epsfig} 
\usepackage{amsmath} 
\usepackage{subcaption}
\usepackage{caption}
\usepackage{color}
\usepackage{verbatim}
\usepackage{wasysym}
\usepackage[table,xcdraw]{xcolor}
\usepackage{graphicx}
\usepackage{gensymb}
\usepackage{xcolor}
\usepackage{latexsym}
\usepackage{multirow}
\usepackage{amsmath}
\usepackage{booktabs,caption,fixltx2e}
\usepackage[flushleft]{threeparttable}
\usepackage[table,xcdraw]{xcolor}
\usepackage{xspace}

\newcommand{\sysname}{REMS\xspace}
\newcommand{\numimp}{32\xspace}

\newcommand{\RNum}[1]{\uppercase\expandafter{\romannumeral #1\relax}}
\makeatletter
\newlength\tmp@\newlength\t@mp
\newcommand{\comp}[3]
  {\mathop{ \settowidth\tmp@{$\displaystyle\mathop{#1}^{#3}_{#2}$}
  \hbox to \tmp@{\hss \settowidth\t@mp{$\displaystyle #1$}\setlength\t@mp{.45\t@mp}
  $\displaystyle\mathop{#1}^{\hspace\t@mp #3}_{\hspace{-\t@mp}#2}$
  \hss} }}
\makeatother
\title{\LARGE \bf
\sysname: Middleware for Robotics Education and Development
}

\author{Yusuke Tanaka$^{\dagger}$, Ankur Mehta$^{\dagger}$
\thanks{$^{\dagger}$ All authors are with the Department of Mechanical and Aerospace Engineering, University of California, Los Angeles, CA, USA 90095 {\tt\small \{yusuketanaka, mehtank\}@ucla.edu.}}}

\begin{document}
\maketitle
\thispagestyle{empty}
\pagestyle{empty}

\begin{abstract}
This paper introduces \sysname, a robotics middleware and control framework that is designed to introduce the Zen of Python to robotics and to improve robotics education and development flow. Although existing middleware can serve hardware abstraction and modularity, setting up environments and learning middleware-specific syntax and procedures are less viable in education. They can curb opportunities to understand robotics concepts, theories, and algorithms. Robotics is a field of integration; students and developers from various backgrounds will be involved in programming. Establishing Pythonic and object-oriented robotic framework in a natural way can enhance modular and abstracted programming for better readability, reusability, and simplicity, but also supports useful and practical skills generally in coding. \sysname is to be a valuable robot educational medium not just as a tool and to be a platform from one robot to multi-agent across hardware, simulation, and analytical model implementations.  

\end{abstract}

\section{INTRODUCTION}

Robotics platforms have been proposed to simplify and generalize programming and integration of robotics technologies across various areas of study. Popular middleware ROS \cite{ros} enables developers to create a package that can run cross-platform with a unified data communication protocol. OROCOS intends to provide a robot control system with a soft real-time capability and built-in robotics libraries. However, such middleware systems require a handful of setup procedures and compiling packages, and the users often should spend efforts to start using them \cite{ros_edu}. This can be a considerable disadvantage when the primary focus is education since students are forced to spare time learning an uncommon language, syntax, and technologies elsewhere. Programming for an integrated system such as a robot can become complex and abstract thinking may not be as intuitive \cite{lego}. LEGO Mindstorms has been a successful tool for elementary to college level classrooms thanks to interactive and fun infrastructure \cite{lego},  \cite{stem}. Although its primary graphical programming makes robot controls more accessible, there is still a gap from LEGO to more advanced or custom robots since Mindstorms is a proprietary system, and all hardware and software are designed to be integrated. 

Python has been gaining popularity because of its simple ecosystem and readability. The Zen of Python is all about simplicity: "Beautiful is better than ugly." and "If the implementation is easy to explain, it may be a good idea." \cite{zen}. 
Introducing the Zen of Python into robotics can philosophically reinforce robotics software and architectures to be better readable and sharable. Adapting Python can benefit users with and without a programming background due to its simple installation mechanisms, interactive no-compilation interpreter, and scientific or machine learning libraries. 
Object-oriented structures benefit and accelerate robotics development with abstracted and modular codes that can be inherited or merged to create a new system. Such scalable and coherent designs have the potential to enhance robotics education and development.



In this paper, we propose \textit{\sysname: Robotics Education Middleware System}, a robotics middleware and control framework that employs the zen of Python and are designed for educational and research purpose.

Our contributions are summarized as follows:

\begin{enumerate}
    \item We develop \sysname: a Pythonic robotics middleware and framework for education and development purpose
    \item Object-oriented design to allow users to swap, extend and combine for a new robot or a functionality
    \item Enable simultaneous run of multiple robots and implementations for a multi-agent system and debugging/comparisons
    \item Case studies running \numimp robot implementations, and applications in an academic class and  research
\end{enumerate}

 \begin{figure}[t!]
    \centering
    \includegraphics[width=0.45\textwidth, trim={0cm 0cm 0cm 0cm},clip]{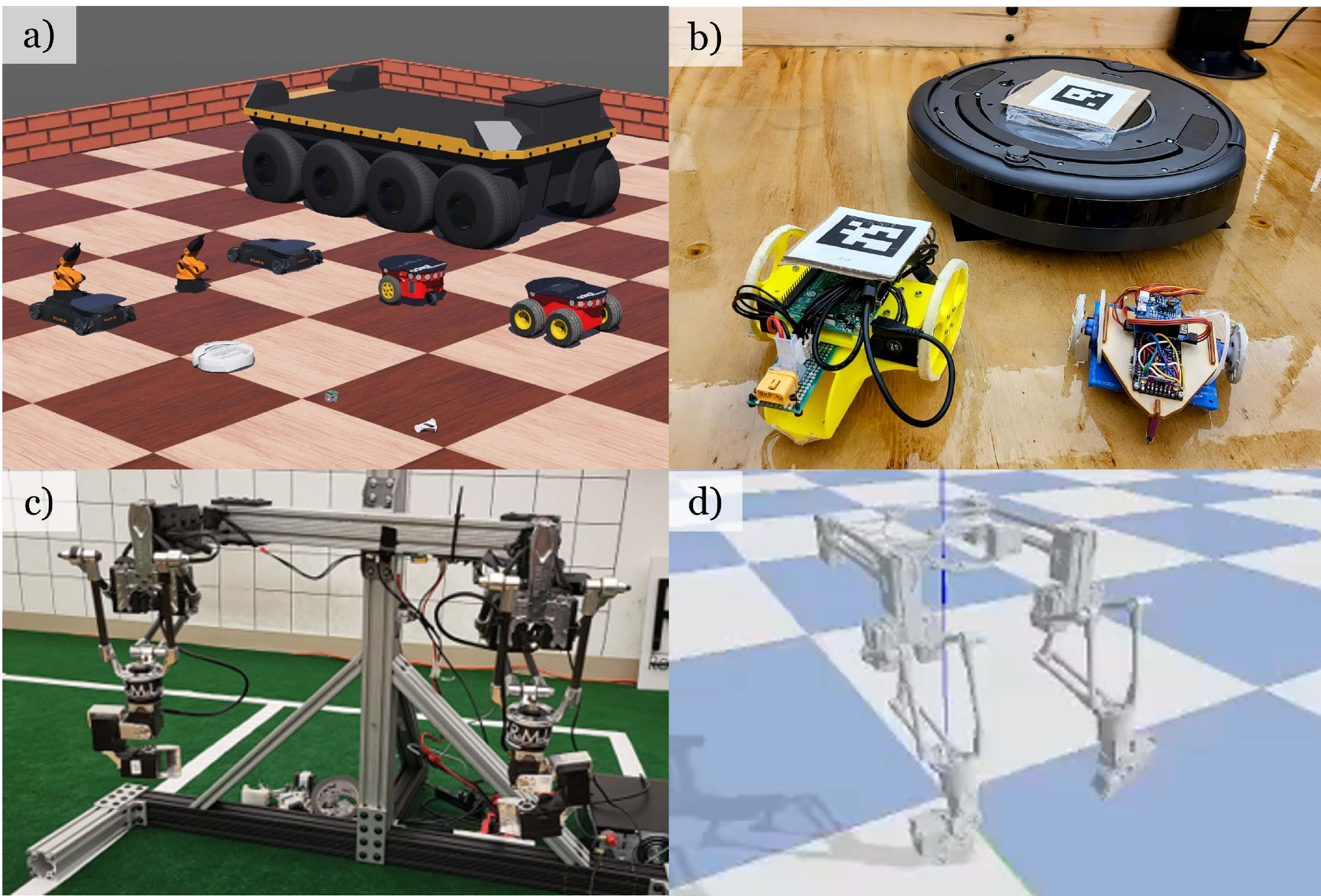}
     \caption{\sysname: robotics middleware and framework. A total of \numimp robot implementations were demonstrated, including simulations in a) and d), and hardware in b) and c), and analytical models. a) Webots physics simulator robots: Woodbot, E-puck, Create 2, Youbot, arm, base only, Pioneer DX, AT, and Moose. b) hardware: Create 2 (top), Dynabot (left) and Woodbot (right). c) Two-arm 6-DoF manipulator hardware, SCALER Manipulator\cite{scaler} and d) Pybullet physics simulation. \label{fig:fig1}}
\end{figure}

\section{Related Works}
\subsection{Middleware in Robotics}
Robot Operation System (ROS and ROS2) is one of the most well-known and widespread middleware for robotics \cite{ros}. ROS provides a hardware abstraction layer (HAL) and allows graph node-edge-based generalized communications. HAL modulates the system's actual implementation, which may depend on specific hardware or programming language, and improves compatibility with other systems and reusability. 
Node connections can be inter-device since ROS 1 uses a custom TCP/IP protocol, and ROS 2 is based on Data Distribution Service (DDS), which is more secure and appropriate for industrial use cases \cite{ros}.
However, ROS itself does not provide robotic system architectures, and the user should design graph communications. Even though a robot is compatible with ROS, it does not mean it is indeed developed using ROS inside, but the ROS node is prepared for a generic external communication medium.

Yet Another Robot Platform (YARP) is an open-sourced middleware to provide HAL and unified communications between modules along with robotics libraries to provide control infrastructure of the robots \cite{yarp}.

In industry, proprietary platforms and programming languages are developed for a particular family of products, such as KUKA Robot Language (KRL) for KUKA robot arms \cite{kuka}, and B\&R Automation Studio for their programmable logic controllers (PLC) \cite{br}. They provide uniform and abstracted access to devices with integrated robotics API libraries and safety systems, although proprietary systems require licensing and specific hardware.

\subsection{Hardware Oriented Middleware}
Open Robot Control Software (OROCOS) is an open-sourced robotic control framework focused on real-time performance \cite{orocos}. Although ROS 2 has enhanced real-time capability, OROCOS has superior latency and jitters in Inter-Process Communication (IPC), particularly with modified Linux kernels such as Xenomai \cite{ros_orocos}. OROCOS framework includes standard APIs for robotics in C++, such as kinematics, Bayesian filtering, state machines, etc.

IPC can restrict the system's real-time operation and cyclic speed since data cannot be referred by memory address, and the data must be copied and serialized to another process \cite{tzc}. \textit{Hard} real-time system guarantee a response time with a certain maximum deviation, while \textit{soft} real-time system bounds average response time \cite{ros_autonomous}. Many attempts toward soft real-time requirements, including OROCOS, have been done in RT-Middleware \cite{rt-middleware} and TZC \cite{tzc}.

Hard real-time is significantly strict and requires low latency hardware, a real-time operating system (RT-OS), and strict time synchronization, primary for safety purpose \cite{ros_autonomous}. The automation industry has developed proprietary PLC to achieve low-latency hard real-time programming environments by custom designing from hardware to RTOS, such as B\&R Automation studio \cite{br}. XBotCore is an open-sourced hard real-time software platform for EtherCAT-based robots \cite{xbot}. 

One of the most successful educational purpose middleware is the LEGO Mindstorms series \cite{lego}. Their Graphical Programming Language (GPL) is suitable for students and beginners unfamiliar with programming. The GPL grants a coding environment as if the users are assembling bricks. For more advanced users, LEGO Mindstorms supports C++ and Python. However, their costs, limited simulation, and proprietary hardware barriers introducing them into classrooms or moving onto different or custom hardware.

\subsection{Middleware for Multi-Agent and Cloud Robotic}
Various middleware has targeted to enhance HAL into a robot, task or team scale to simplify programming at different levels and modulate them to reuse. Task-level GPLs, RAZER, etc., are toward end-users so they can program a robot during operations and without an engineer \cite{razer}. They can help facilitate automation and robotics technologies to lesser-technical creative industries with human-robot interaction in mind \cite{creative}. ManiWare is dedicated to team-level abstraction and synchronization for manipulators to conduct teamwork operations among different arms \cite{maniware}. CORNET integrates unmanned aerial vehicle physics and network simulators to realize a cyber-physical system of multi-agent drones \cite{cornet}. 

Cloud robotics offers offloading computation-intensive tasks and access to cloud resources such as CPU, GPU, storage, and a global map \cite{cloud_robotics}. Multi-agent and collaborative robots can benefit from cloud infrastructure. XbotCloud is a cloud-integrated system designed for XbotCore-operated robots \cite{xbotcloud}.

\section{\sysname Concept and Principle}
\subsection{Concepts}
Our middleware , \sysname, design primary concepts are the following:
\begin{itemize}
    \item The Zen of Python to robotics
    \item Provide robotic framework off-the-shelve
    \item Moduled robot definitions and implementations
    \item Scalable from one to multi-agent
\end{itemize}

The concept of Pythonic programming is about simple, concise, and readable coding. Each class and method have a clear role in the system. With other common middleware, learning curves are steep at the beginning because users should learn the specific syntax, language, process, etc. of the middleware that are not standard or uncommon elsewhere, i.e., IDL in ROS. Even setting up and installing a ROS package is already a significant achievement due to complicated version dependencies \cite{ros_edu}. The robotics domain spreads not merely to computer science but to people with a mechanical or electrical engineering background. In education settings, it can take away vital opportunities to learn more about theories and algorithms by spending time on technical details that are only applicable in a limited field area. 

Developers should put considerable design efforts into software architecture to integrate various components in robotics. Providing a robot control pipeline simplifies the initial setup and implementation time. The architecture itself can educate learners on what parts should be in the robot system as well as basic concepts of programming, such as abstraction and modularity. 

Abstraction in a robot is more complex than said. What defines a robot to be and what does not, are not clear or less considered questions. Coherent divisions between a robot definition and its actual implementation support students' better understanding and execution of the theories lectured. This strategy lets users swap the implementation from hardware to simulation or to an analytical mathematical model of the same robot, which can help them debug their math or algorithms by cross-checking the outcomes and can prevent damaging money and time extensive hardware. 
Users can inherit, extend and merge one or more existing definitions to create a new robot definition, which expands the possibility of rapid development and modular robot software design.
Moving from one robot to two or multi-agent is not trivial since there must be another layer of software designed to handle and communicate with all the robots. Seamless scalability of more robots remove a barrier to getting started with a multi-agent system; moreover, users can add the same robot with different implementations. 

With these concepts, \sysname ambitions to be a valuable educational medium not just as a tool but as a platform that can support advanced robotics development and research. 

\subsection{Design Principle}

\begin{itemize}
    \item Abstraction and modularity
    \item Generalize interface and conversions
    \item Processing, threading, 
    \item synchronization and synchronization
\end{itemize} 

\begin{figure}[h!]
    \centering
    \includegraphics[width=0.45\textwidth, trim={0.5cm 0cm 0cm 0cm},clip]{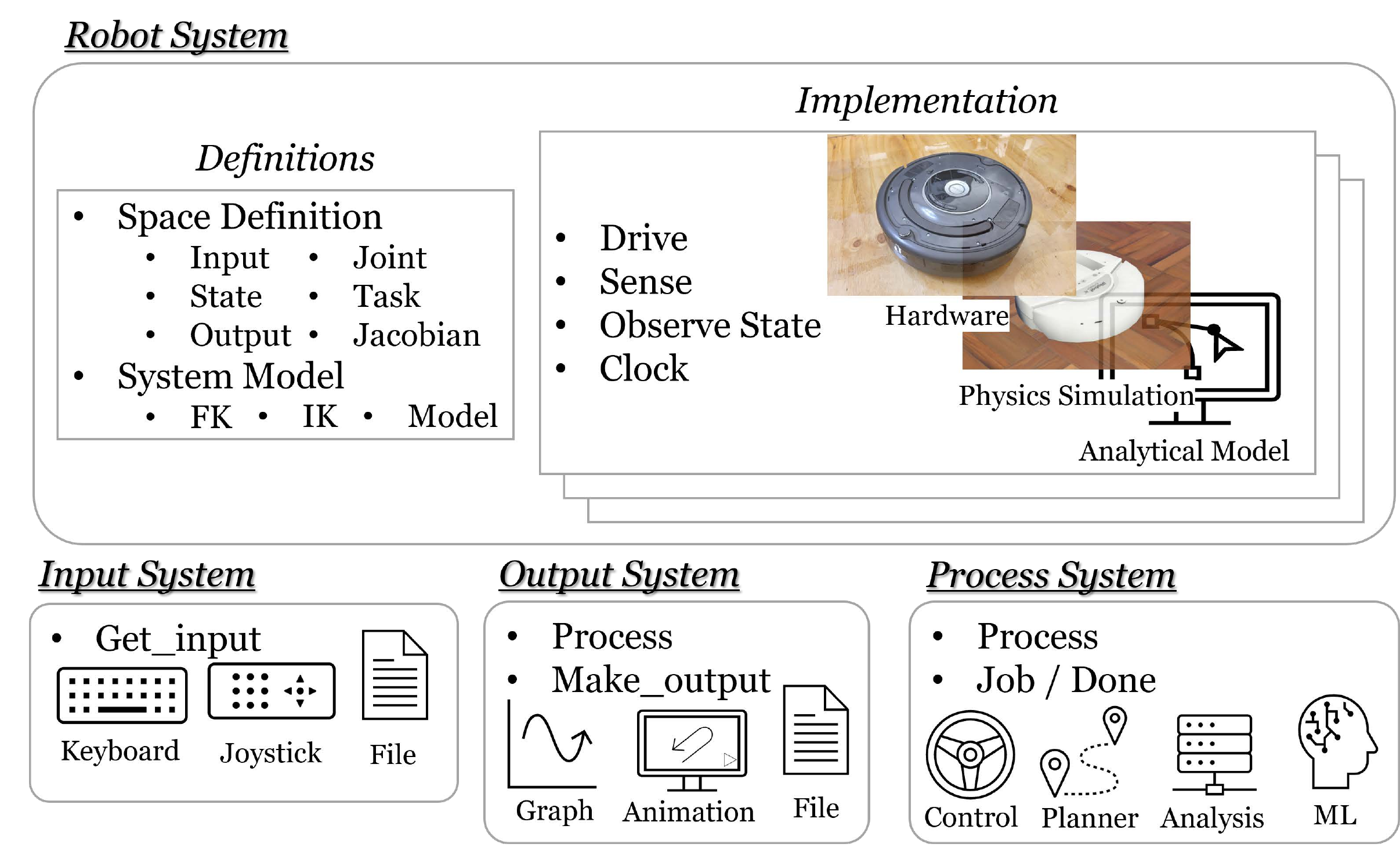}
     \caption{\sysname middleware system class, variables and method list. 
     The robot system consists of a definition and implementation. Definitions contain the robot's unique information. Implementation represents hardware, simulation, or analytical model and how to interact with them. 
     This facilitates an easily swappable mechanism for different robot definitions and implementations. \label{fig:system}}
\end{figure}

Abstraction and modularity are the basics of programming. Clarifying the dependency of each module can enhance its reusability. In \sysname, we divide a whole robot into a set of \textit{sub-system} as shown in \fig{fig:system}. Each system should inherit from a corresponding base and be independent of other systems. This approach resolves the complex integration of a robot structure into a manageable scale and encourages users to separate the functionality they want to achieve into independent components. Each system, class, and method is designed to do an explicit functionality that can be read from their names.

Interfaces and communications among different modules must be abstracted to ensure each system can adequately communicate. Common methods, such as having structured data classes or positional assignments using array objects, are prone to change and not explicit or Pythonic. \sysname aims to attain an explicit but flexible and natural way of interfacing. However, it is also important to consider cases when not a standard form of variables is assigned since Python is a dynamically typed language and type hints are not mandatory. 

Python processing and threading are unique due to Global Interpreter Lock (GIL). GIL limits each python interpreter to one CPU core, and no concurrent thread execution is allowed. It can quickly handicap a python-based robot framework, especially when adding multiple robots, graphical outputs, or any computationally extensive processes. 
Many popular libraries and simulators have restrictions on using them in a separate thread or having multiple instances inside the same processes. To evade those restrictions in Python, multi-process-based robot execution is more appropriate. Multi-processing can increase architecture design and coding complexity, and IPC handling is not necessarily intuitive for all users. We design \sysname to be as natural as possible and ensure that such multi-processing issues do not arise and cause minimal pain to the users. 

Synchronous and asynchronous executions become essential, especially when hardware devices are involved. It is not reasonable to slow down the entire system because one of the devices is slow to extract data. However, asynchronous operations make the data transmission non-deterministic, and some communication protocol requires ordered connections. Half-duplex serial links only allow one-way communication simultaneously. Thus it is more appropriate to sequentially write and read to such devices, i.g. actuator commands and encoder readings. For I/O bounded operations, threading can be adequate to avoid IPC. Consequently, \sysname accommodates users’ needs by supporting several types of executions mode.

\section{Middleware Architecture\label{sec:architecture}}

\subsection{Input, Robot and Output System}
The user should specify three systems to run a robot: input, robot, and output system. An input system takes inputs to the robots, such as a trajectory file, keyboard, or joystick. The robot system contains robot definitions and implementations necessary to interact with the robot. An output system handles how to save or display data collected at each timestep and after the run.

\subsection{Robot Definitions and Implementations}
The robot system separates robot definitions and implementations. The Definitions include the robot space definitions and forward, inverse kinematics, etc., which are universal regardless of the actual implementation of the robot. The implementation means such as simulation or hardware. Dividing a robot definition class and an implementation class explicitly forces the users to think about what defines a robot to be and what is tight to a specific simulation or hardware. This improves the reusability of such definitions and implementations. For instance, \sysname consists of standard support for a Webots psychics simulator. Thus, users can apply their robot definitions to the Webots implementation to use simulations with little extra coding.
When adding a robot to a run, users hand over these two, and under the hood, they are dynamically inherited to create a one single robot class instance. Therefore, more advanced users can overwrite robot implementation behaviors in the definition class, or vice versa, as the instantiated robot is derived from these two (Python multiple inheritances).

\subsection{Interfacing, Data Flow and Conversions}
Interfacing among different codes can easily cause issues when a data mismatch happens. 
In \sysname, data are stored in a dictionary-like object; we call it \textit{Defined Dictionary} or DefDict. DefDict shares similar ideas to structured array in NumPy, which names each array-like element for more explicit data management. 
DefDict is designed to work as if it is a Python dictionary object and holds set of variable names and corresponding element types. 
Primary differences from the existing libraries are:
\begin{itemize}
    \item Type enforcement and optional unit enforcement
    \item Add rules for conversions among different definitions
    \item Object-oriented nested data structure handling
\end{itemize}
Instead of specifying Python types such as float and int, users can set the data unit, range, mapping tables, default, etc., with our custom unit class. Basic unit conversions and dimension checks take advantage of Unyt library \cite{unyt}. Unconventional translation can occur when i.g. an actuator takes a duty cycle (PWM), or a sensor returns byte data that needs to be solved by a lookup table. 
Users can add a custom rule to define a mapping, such as from keyboard commands to robot inputs.   


\subsection{Process System and Background Job}
The processing system get executed at each system timestep. It helps run data processing, such as mapping, online planning, or machine learning. The user can interact with robot classes by handing over the instance of the robots at setup. The processing system can dispatch background jobs and receive a callback when the job is done. Time-consuming computations that may take longer than one timestep can benefit, such as image processing.

 \begin{figure}[h!]
    \centering
    \includegraphics[width=0.35\textwidth, trim={0cm 0cm 0cm 0cm},clip]{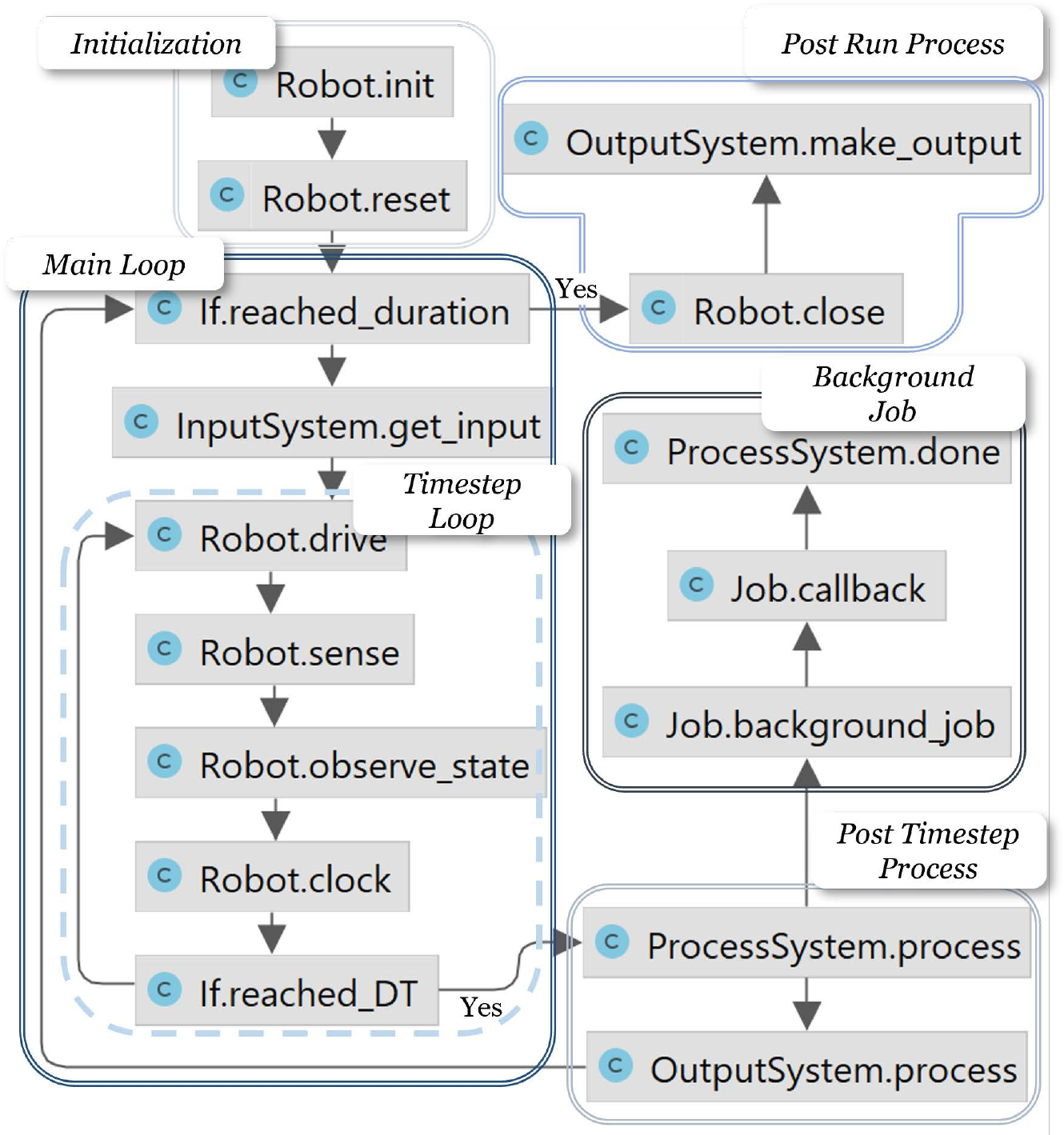}
     \caption{Robot control framework execution orders. After initialization, the main loop will be executed at a system timestep till the user-specified time. After the run, the robot is closed, and outputs are generated. Process system and output system are called at every system timestep. The background job sends a callback to the submitter after the job is done. \label{fig:architecture}}
\end{figure} 

\subsubsection{Time Synchronization, Real-time and Inter/Intra Process Communication}
Time synchronization and real-time are often issues in robotics. However, real-time system demands specially designed hardware and operation system (OS) or kernel. 
Therefore, strict time management is out of our scope because our focus is Pythonic installation and simple usage for everyone. 
\sysname time synchronizes using a discrete timestep, and all robots receive the same time from the primary process. Systems, robots, processes, and devices can have different internal timestep, which is handled in an inner loop as shown in \fig{fig:architecture}.
Optionally, our system lets users run robots faster than in actual time or as fast as possible if they are in simulations. 

\sysname run independent processing for each robot and output using a Ray library \cite{ray}. 
Devices runs on thread and optionally in a different process. 
\sysname hides Ray API 
and emulates behaviors of ordinary Python objects so that users can handle them as naturally as possible. 

\section{Case Studies and Applications}
 \begin{figure}[h!]
    \centering
    \includegraphics[width=0.49\textwidth, trim={0cm 0cm 0cm 0cm},clip]{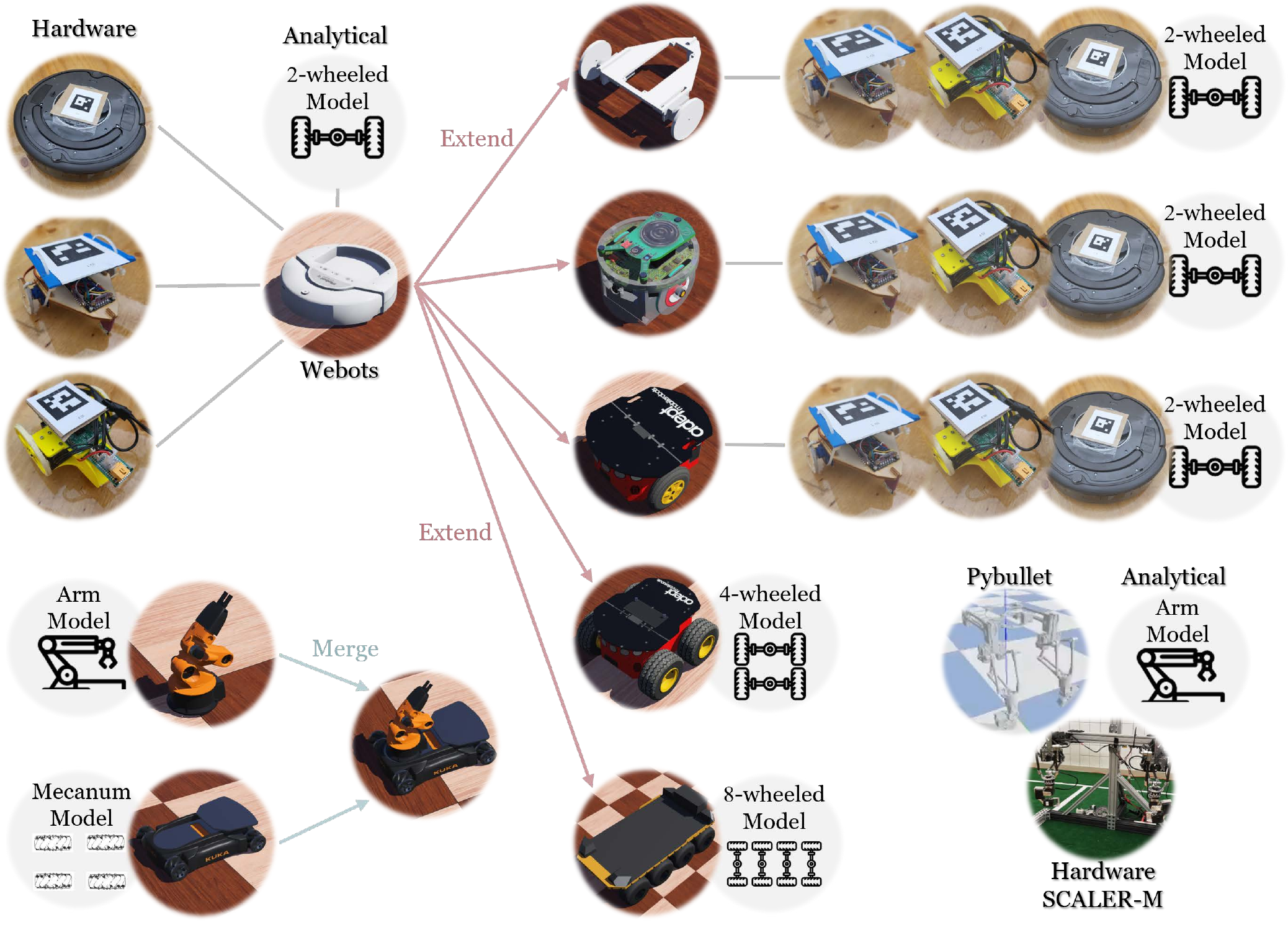}
     \caption{\sysname hardware, simulation and analytical model experimented in the case studies. Total \numimp implementations were pictured. All differential drive robots are inherited and extended from the same base. Youbot is defined by merely merging base and arm definitions. \label{fig:implementation}}
\end{figure} 
We have conducted the following case studies to exhibit our middleware system capabilities, and applications enabled. Robots implemented are pictured in \fig{fig:implementation}.
\begin{enumerate}
    \item Middleware use examples with various robot definitions, types, and implementations
    \item Educational uses in an undergraduate academic class 
    \item Academic research with a high DoF robot 
    \item Multi-agent cases with different implementations
\end{enumerate}

\subsection{Webots Examples \label{webots_examples}}
We use Webots examples of robots to show how our system can easily control different robots implemented in kinematics, simulation, and hardware. Differential-drive, omni-drive, and manipulator systems were selected. A manipulator use case is also discussed in Sec. \ref{real2sim}. Three types of drive mechanisms were implemented in Webots simulation and analytical modeling. One robot was tested with three different hardware. 

\begin{itemize}
    \item Create 2, iRobot, (Two-wheeled)
    \item Youbot base, Kuka, (Mecanum Wheeled, Omni drive)
    \item KUKA manipulator, Kuka, (5-DoF manipulator)
\end{itemize}

Create 2 was experimented with Create 2 hardware and two custom-designed two-wheeled robots; we call them; Woodbot and Dynabot.
Create 2 is an off-the-shelf Roomba-like robot with touch and distance sensors tethered through USB serial communication.
Woodbot is an affordable differential drive robot generated through \cite{RoCo} and manufactured using a laser cutter. A microcontroller, ESP32, controls Woodbot with a 9g continuous servo, two lidar, and IMU sensors, and it can communicate to a computer through a Wi-Fi WebSocket connection. 
Dynabot is a 3D-printed robot using Dynamixel actuators and tethered.
Arauco markers were attached for their state estimations. 

Interfacing issues in the codes are not negligible in this case since the robot definitions are used with five implementations: analytical model, Webots, and three hardware. Some implementations use different units, and they all work distinctly.
For instance, actuator velocity in Webots is in rad/s, Dynabot in rpm, Woodbot in duty cycle, and Create 2 in count range $[-500, 500]$. Such interface discrepancies are automatically handled in DefDict with the unit system. 

\sysname main loop can run at relatively higher frequency (i.g. 1kHz), but typically hardware communication limits the maximum device frequency. Create 2 is at 20Hz, Dynabot at 100Hz, Woodbot at 5Hz for sending inputs and receiving sensor outputs. Timestep loop and device threads handle the frequency difference asynchronously.  

\subsection{Extending Existing Robots for Different Definitions}
The previous case demonstrated that \sysname can adopt three types of robots with various implementations. We can further inherit those robot definitions to generate and accommodate similar but different robot definitions. We extended three robots definitions to test the following robots:   

\begin{itemize}
    \item Woodbot, (Two-wheeled)
    \item E-puck, GCTronic, (Two-wheeled)
    \item Pioneer 3DX, Adept, (Two-wheeled)
    \item Pioneer 3AT, Adept, (Four-wheeled)
    \item Moose, Clearpath Robotics, (Eight-wheeled)
    \item Youbot with an arm, Kuka, (Mecanum Wheeled, omni drive with a 6-DoF manipulator)
\end{itemize}

Create 2, Woodbot, E-puck, Pioneer 3-DX, AT, and Moose are all differential drive robots, meaning they all can be controlled in the same manner. Therefore they can all inherit the same differential drive definition class. However, Pioneer 3-At and Moose include slave motors. Such a case was resolved by adding rules to link the master and slave motors. 
Youbot installed with a KUKA manipulator is a combination of the two earlier examples, meaning we can simply inherit two definitions as they don't have any interfering definitions. 

The two-wheel robots are also tested on three hardware as well. Although the manipulator case is not tested in hardware here, Sec. \ref{real2sim} demonstrated that two-arm 6-DoF manipulator with Analytical model, simulation, and hardware. 


\subsection{Woodbot Class Lab Activity}
 \begin{figure}[h!]
    \centering
    \includegraphics[width=0.49\textwidth, trim={0cm 0cm 0cm 0cm},clip]{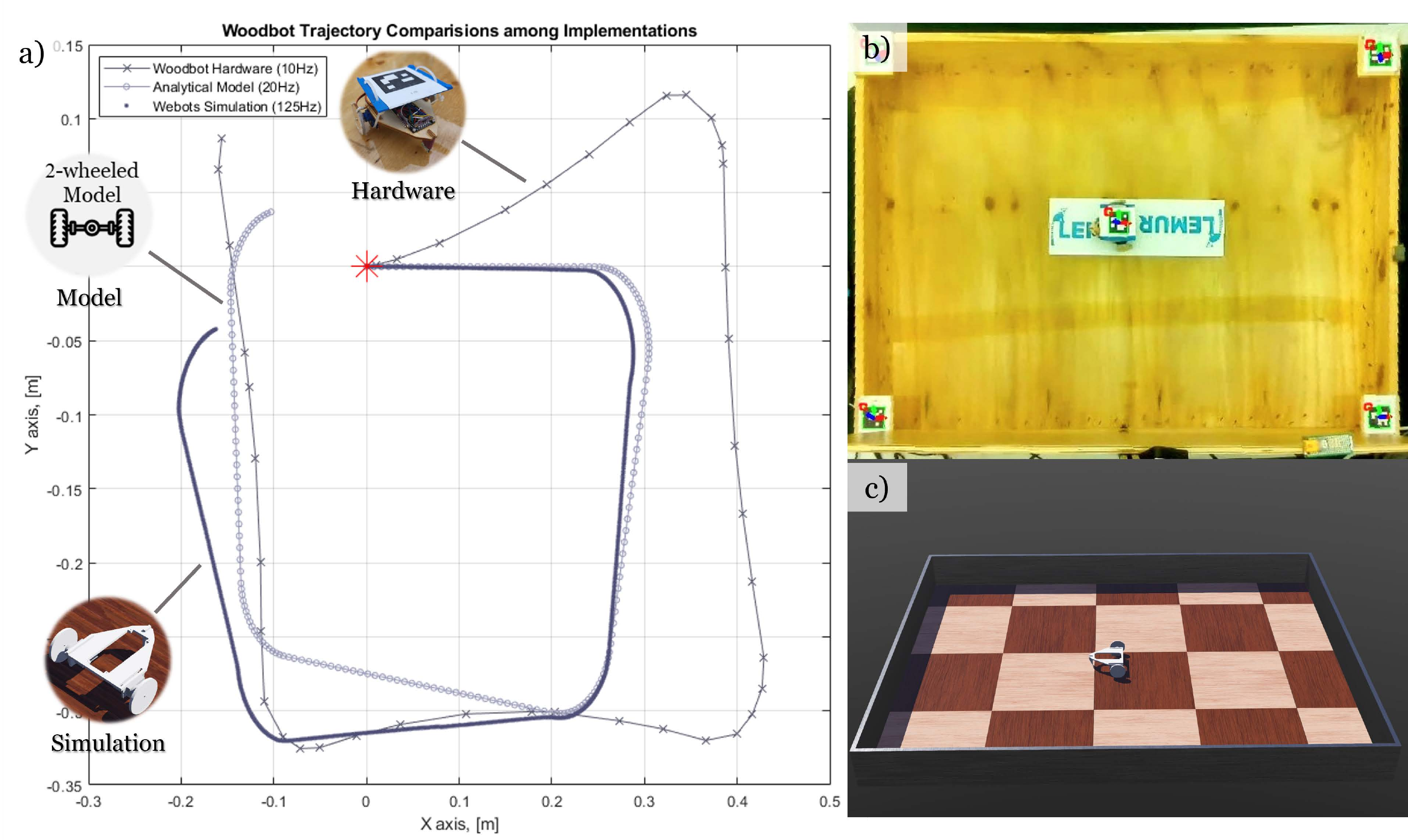}
     \caption{Woodbot trajectory comparison. Students were tasked to formulate kinematics model of Woodbot, setup simulation, assemble hardware, and run them all to compare. a) Three implementations (hardware, simulation, model) simultaneously ran with the same input starting from the red star for $10$ s. b) Hardware robot and c) Webots simulation environment. \label{fig:woodbot}}
\end{figure} 
We experimented with our earlier version of our middleware in a robotics capstone class, an undergraduate student graduation project course for the Electrical and Computer Engineering department and Mechanical and Aerospace department (ECE183, MAE162) at UCLA. 
The groups of students were tasked to derive a kinematics model of Woodbot motion and sensor outputs, implement a Webots simulation, and assemble and run Woodbot hardware shown in \fig{fig:woodbot}. These tasks were to familiarize them with basic robotics knowledge and to prepare them for their own robotics final project. 
Our middleware lets them run their derived kinematics, simulation, and hardware using human inputs or predefined trajectories and visualize their motions and states. Running three implementations simultaneously demonstrated that these motions were generally close but also not identical. In \fig{fig:woodbot} example, the Webots simulation and analytical model behaved closely at the beginning and slightly diverged at the end. Hardware tended to lean left when going straight. 
This helped students debug their kinematics model calculation by comparing the states and outputs in real-time and offline. 
Students have learned what fundamentally defines the robot theoretically and what it indeed takes to run robots.

\subsection{Real to Sim Dynamics Residual Modeling \label{real2sim}}
 \begin{figure}[h!]
    \centering
    \includegraphics[width=0.49\textwidth, trim={0cm 0cm 0cm 0cm},clip]{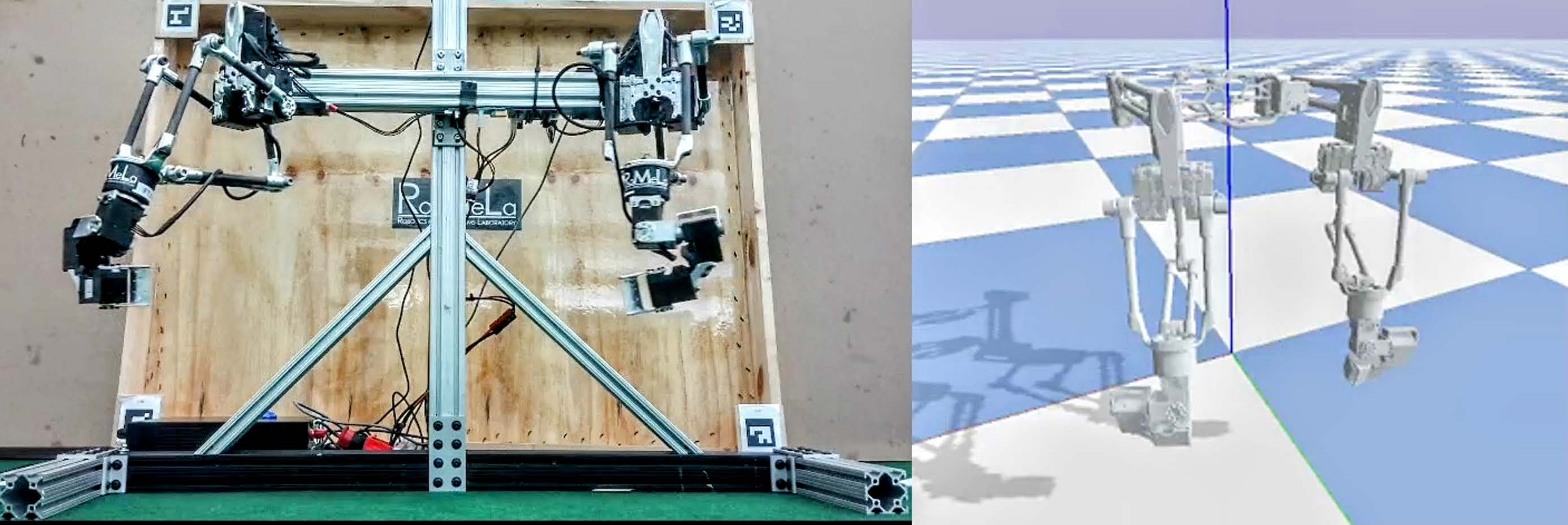}
     \caption{SCALER-M, a two-arm 6-DoF Manipulator in hardware and simulation. SCALER-M is a manipulator configuration of SCALER: a quadruped climbing robot in \cite{scaler}. Analytical model is based on forward kinematics.  \label{fig:scaler_manipulator}}
\end{figure} 

The residual dynamics modeling attempts to capture the discrepancy between simulation and hardware behaviors. This requires the same robot in both simulation and hardware to collect the data. \sysname is helpful in this case since the users can add the same robot with different implementations and run them simultaneously with the same inputs. 
In this application \cite{real2sim}, three residual models were generated for a mobile and a stationary robot system: between kinematics model and hardware, between kinematics model and simulation, and between simulations and hardware. 
Woodbot and a two-arm manipulator, SCALER manipulator configuration \cite{scaler} with 6-DoF per arm were used.

Deep neural network (DNN) was back-propagated online using an unscented Kalman filter-based (UKF) auto-tuner. 
This UKF auto-tuner can run online and update the residual model, which provides strength to the pipeline. Therefore, \sysname not only allows training the residual models but also helps to keep updating these models online while the users run their trajectories. A process system was used to collect and handle data from two robot implementations. DNN back-propagation was done as background jobs since it was computationally extensive and was not time critical. 

The models’ quality was compared and evaluated by running the robots simultaneously with and without a residual term, such as by running two simulation instances or treating each arm as an individual robot. The SCALER manipulator simulation was experimentally done in Pybullet physics simulator for Python. 

\subsection{Multi-Agent}
\sysname is designed to run multiple robots simultaneously regardless of whether they share the same robot definitions or implementations. 
We placed seven Webots simulation example robots and three hardware in Sec. \ref{webots_examples} as in \fig{fig:fig1} a), b). They were operated either: using a system-wide input to control all the robots or with individual robots independent inputs, which mimic global and local controls structure in multi-agent systems.

\begin{table}[]
\begin{threeparttable}
\caption{List of all robots and implementations experimented with.}
\begin{tabular}{lccc}
\rowcolor[HTML]{ECF4FF} 
Robot & Anaytical & Simulation & Hardware \\ \hline
Create 2 & Jacobian & Webots & Create, Dynabot, Woodbot \\
\rowcolor[HTML]{EFEFEF} 
Woodbot & Jacobian & Webots & Create, Dynabot, Woodbot \\
E-puck & Jacobian & Webots & Create, Dynabot, Woodbot \\
\rowcolor[HTML]{EFEFEF} 
Pioneer 3 DX & Jacobian & Webots & Create, Dynabot, Woodbot \\ \hline
Pioneer 3 AX & Jacobian & Webots &  \\
\rowcolor[HTML]{EFEFEF} 
Moose & Jacobian & Webots &  \\ \hline
Youbot base & Jacobian & Webots &  \\
Youbot & Jacob/FK & Webots & \multicolumn{1}{l}{} \\ \hline
\rowcolor[HTML]{EFEFEF} 
KUKA Arm & FK & Webots &  \\
SCALER-M & FK & Pybullet & Two SCALER Manipulator \\ \hline
\end{tabular}

\begin{tablenotes}
      \small
      \item Total \numimp of robot implementations experimented with throughout our case studies for ten different robot definitions.
    \end{tablenotes}
\end{threeparttable}
\end{table}

\section{Discussion}
Our case studies demonstrated that our middleware has simplified and modularized the robotics coding flow, allowing students and researchers to focus on what they want to explore. 
Wheeled robot examples exhibit how one robot's definition can be extended to adapt similar robots. Combining an omni-drive base and arm definitions, we realized a complete assembly of Youbot control. This object-oriented way can not only simplify defining a new robot but also reduce maintenance labor since a fix in a base will be applied to all the robots inherited from it.

We tested differential drive robots on three different two-wheeled hardware implementations. This capability means that the same control codes can be tested on different sets of actuators and sensors. Since Woodbot is made out of very affordable components, the actuator velocity control is less accurate, making it harder to move straight than the other. 

Swapping implementations with the same robot definition enabled students to quickly test a robot on hardware, simulation, and analytical model and then compare and validate the results. \sysname enhanced the students' experience by minimizing implementation works and providing self-validation methods. 

In the academic research case, \sysname aided and added an edge to the residual modeling algorithm. The algorithm can run and improve the model online since \sysname lets users run hardware and simulation efficiently. 

\sysname has been applied from wheeled to high-DoF manipulator robots, from student-derived models to Webots, Pybullet simulations, to off-the-shelf hardware and embedded system. Applications are not limited to a simple joystick control or an offline trajectory, but it is capable of machine learning and supporting multi-agent systems. Furthermore, it can enable multi-agent research in mixed reality, where robots collaborate, for example, in the real and virtual simulation worlds.

\section{Conclusion and Future works}
We have developed \sysname that introduces the Zen of Python to robotics to be more accessible and educational. Our object-oriented framework designs enhance modularity both theoretically and practically. We showcased how various robot types and implementations can be executed interchangeably and how users can inherit and merge existing robots to create new ones. Case studies in an academic class and academic research demonstrated that \sysname revamped and accelerated robot education and research. 

Now \sysname opens up a wide range of possibilities such as sim-to-real translation in the same robot definition or among extended or merged definitions, a quick algorithms performance benchmark with diverse robots, or a multi-agent system across hardware and simulation.


\begin{thebibliography}{99}

\bibitem{ros} S. Macenski, et al. "Robot Operating System 2: Design, architecture, and uses in the wild," \emph{Science Robotics}, vol. 7, no. 66, pp. eabm6074, 2022.

\bibitem{ros_edu} K. Hauser, "Why I Don’t Teach ROS to Robotics Students," Medium, Online, 2021. https://hauser-kris.medium.com/why-i-dont-teach-ros-to-robotics-students-55cc4f3623ce

\bibitem{lego} S. R. Perez, C. Gold-Veerkamp, J. Abke and K. Borgeest, "A new didactic method for programming in C for freshmen students using LEGO mindstorms EV3,"  \emph{International Conference on Interactive Collaborative Learning (ICL),} pp. 911-914, 2015.

\bibitem{stem} E. Afari, and S. K. Myint, "Robotics as an educational tool: Impact of lego mindstorms." \emph{International Journal of Information and Education Technology 7.6):} 437-442, 2017.

\bibitem{zen} T. Peters, "The zen of python." Pro Python. Apress, 301-302, 2010.

 
\bibitem{yarp} G. Metta, et al. “YARP: Yet Another Robot Platform.” \emph{International Journal of Advanced Robotic Systems,} 2006.



\bibitem{kuka} H. Muhe, A. Angerer, A. Hoffmann, and W. Reif, "On reverse-engineering the KUKA Robot Language, " arXiv:1009.5004, 2010.

\bibitem{br} "B\&R Industrial Automation GmbH: Automation Studio," 2022. https://www.br-automation.com/en-us/products/software/automation-software/automation-studio/

\bibitem{orocos} H. Bruyninckx, "Open robot control software: the OROCOS project,"\emph{Proceedings ICRA. IEEE International Conference on Robotics and Automation,} pp. 2523-2528 vol.3, 2001.

\bibitem{ros_orocos} S. Barut, M. Boneberger, P. Mohammadi and J. J. Steil, "Benchmarking Real-Time Capabilities of ROS 2 and OROCOS for Robotics Applications," \emph{IEEE International Conference on Robotics and Automation (ICRA),} pp. 708-714, 2021.

\bibitem{tzc} Y. P. Wang, et al. "TZC: Efficient Inter-Process Communication for Robotics Middleware with Partial Serialization,"  \emph{IEEE/RSJ International Conference on Intelligent Robots and Systems (IROS),} pp. 7805-7812, 2019.

\bibitem{ros_autonomous} A. M. Hellmund, et al., "Robot operating system: A modular software framework for automated driving," \emph{IEEE 19th International Conference on Intelligent Transportation Systems (ITSC)}, pp. 1564-1570, 2016.

\bibitem{rt-middleware} N. Ando, et al., "RT-middleware: distributed component middleware for RT (robot technology)," \emph{IEEE/RSJ International Conference on Intelligent Robots and Systems (IROS),} , pp. 3933-3938, 2005.

\bibitem{xbot} L. Muratore, et al., "XBotCore: A Real-Time Cross-Robot Software Platform," \emph{First IEEE International Conference on Robotic Computing (IRC),} pp. 77-80, 2017.

\bibitem{razer} F. Steinmetz, A. Wollschläger and R. Weitschat, "RAZER—A HRI for Visual Task-Level Programming and Intuitive Skill Parameterization," \emph{ in IEEE Robotics and Automation Letters,} vol. 3, no. 3, pp. 1362-1369, 2018.


\bibitem{creative} J. Braumann and K. Singline, "Towards Real-Time Interaction with Industrial Robots in the Creative Industries," \emph{IEEE International Conference on Robotics and Automation (ICRA),} , pp. 9453-9459, 2021.

\bibitem{maniware} Z. Cheng, J. Cao and J. Chen, "ManiWare: An Easy-to-Use Middleware for Cooperative Manipulator Teams," \emph{IEEE International Conference on Smart Computing (SMARTCOMP),} pp. 349-355, 2022.

\bibitem{cornet} S. Acharya, et al., "CORNET: A Co-Simulation Middleware for Robot Networks,"\emph{ International Conference on COMmunication Systems and NETworkS (COMSNETS),} pp. 245-251,2020. 

\bibitem{cloud_robotics} O. Saha, and D. Prithviraj, "A comprehensive survey of recent trends in cloud robotics architectures and applications." \emph{Robotics} vol. 7 no. 3:47. 2018.

\bibitem{xbotcloud} L. Muratore, B. Lennox and N. G. Tsagarakis, "XBotCloud: A Scalable Cloud Computing Infrastructure for XBot Powered Robots," \emph{IEEE/RSJ International Conference on Intelligent Robots and Systems (IROS), }pp. 1-9, 2018.

\bibitem{unyt} Goldbaum et al., "unyt: Handle, manipulate, and convert data with units in Python," \emph{Journal of Open Source Software,} 3(28), 809, 2018.

\bibitem{ray} Moritz, Philipp, et al. "Ray: A distributed framework for emerging {AI} applications." \emph{13th USENIX Symposium on Operating Systems Design and Implementation (OSDI 18),} 2018.

\bibitem{RoCo} A. M. Mehta, J. DelPreto, B. Shaya and D. Rus, "Cogeneration of mechanical, electrical, and software designs for printable robots from structural specifications,"  \emph{2014 IEEE/RSJ International Conference on Intelligent Robots and Systems,} pp. 2892-2897, 2014.

\bibitem{real2sim} A. Schperberg, Y. Tanaka, F. Xu, M. Menner, D. Hong "Real-to-Sim: Deep Learning with Auto-Tuning to Predict Residual Errors using Sparse Data" arXiv:2209.03210, 2022

\bibitem{scaler} Y. Tanaka et al., “SCALER: A tough versatile quadruped free-climber robot,” \emph{in Proc. 2022 IEEE/RSJ Int. Conf. Intell. Rob. Syst. (IROS), 2022.}


\end{thebibliography}
\end{document}